\documentclass[runningheads]{llncs}
\usepackage[T1]{fontenc}
\usepackage{graphicx}
\usepackage{amssymb}
\usepackage{subfigure}
\newcommand\ie{\emph{i.e.}} 
\newcommand{\etal}{\textit{et al.}}
\begin{document}
\title{EPVT: Environment-aware Prompt Vision Transformer for Domain Generalization in Skin Lesion Recognition}
\author{Siyuan Yan\inst{1,2}, Chi Liu\inst{2}, Zhen Yu\inst{2}, Lie Ju\inst{2}, Dwarikanath Mahapatra\inst{5}, 
Victoria Mar\inst{3}, Monika Janda\inst{4}, Peter Soyer\inst{4}
Zongyuan Ge \inst{2}
}

\institute{Faculty of Engineering, Monash University, Melbourne, Australia \and
Monash Medical AI Group, Monash University, Melbourne, Australia \and
Victorian Melanoma Service, Alfred Health, Melbourne, Australia \and
The University of Queensland Diamantina Institute, Dermatology Research Centre, \\The University of Queensland, Brisbane, Australia \and
Inception Institute of AI, Abu Dhabi, UAE\\
}

%
\maketitle              
\begin{abstract}


Skin lesion recognition using deep learning has made remarkable progress, and there is an increasing need for deploying these systems in real-world scenarios. However, recent research has revealed that deep neural networks for skin lesion recognition may overly depend on disease-irrelevant image artifacts (\ie~dark corners, dense hairs), leading to poor generalization in unseen environments. To address this issue, we propose a novel domain generalization method called EPVT, which involves embedding prompts into the vision transformer to collaboratively learn knowledge from diverse domains. Concretely, EPVT leverages a set of domain prompts, each of which plays as a domain expert, to capture domain-specific knowledge; and a shared prompt for general knowledge over the entire dataset. To facilitate knowledge sharing and the interaction of different prompts, we introduce a domain prompt generator that enables low-rank multiplicative updates between domain prompts and the shared prompt. A domain mixup strategy is additionally devised to reduce the co-occurring artifacts in each domain, which allows for more flexible decision margins and mitigates the issue of incorrectly assigned domain labels. Experiments on four out-of-distribution datasets and six different biased ISIC datasets demonstrate the superior generalization ability of EPVT in skin lesion recognition across various environments. Code is avaliable at https://github.com/SiyuanYan1/EPVT.

\end{abstract}

\section{Introduction}

Skin cancer is a serious and widespread form of cancer that requires early detection for successful treatment. Computer-aided diagnosis systems (CAD) using deep learning models have shown promise in accurate and efficient skin lesion diagnosis. However, recent research has revealed that the success of these models may be a result of overly relying on "spurious cues" in dermoscopic images, such as rulers, gel bubbles, dark corners, and hairs \cite{artifacts,debias1,debias2,trustderm}, which leads to unreliable diagnoses. When a deep learning model overfits specific artifacts instead of learning the correct dermoscopic patterns, it may fail to identify skin lesions in real-world environments where the artifacts are absent or inconsistent. 

\begin{figure*}[!t]
   \begin{center}
   {\includegraphics[width=0.9\linewidth]{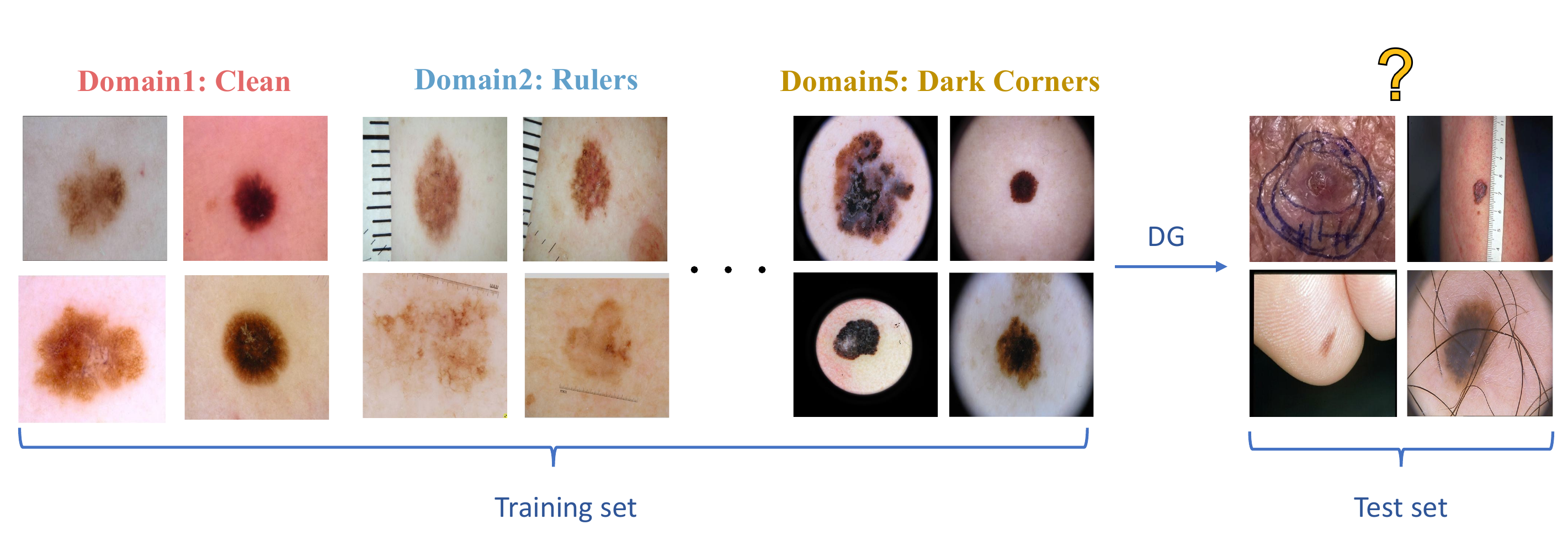}}
   \end{center}
  \vspace{-5mm}
\caption{\small 
The training data is split into five domains: clean, rulers, hairs, air pockets, and dark corners. Domain generalization aims to train the model to learn from these domains to generalize well in unseen domains.}
   \label{fig:motivation}
\end{figure*}

To alleviate the artifact bias and enhance the model's generalization ability, we rethink the problem from the domain generalization (DG) perspective, where a model trained within multiple different but related domains are expected to perform well in unseen test domains. As illustrated in Figure \ref{fig:motivation}, we define the domain labels based on the types of artifacts present in the training images, which can provide environment-aware prior knowledge reflecting a range of noisy contexts. By doing this, we can develop a DG algorithm to learn the generalized and robust features from diverse domains.


Previous DG algorithms learning domain-invariant features from source domains have succeeded in natural image tasks \cite{IRM,inv2,inv1}, but cannot directly apply to medical images, in particular skin images, due to the vast cross-domain diversity of skin lesions in terms of shapes, colors, textures, etc. As each domain contains ad hoc intrinsic knowledge, learning domain-invariant features is highly challenging. One promising way is, as suggested in some recent works \cite{reid,norm,domain_adap}, exploiting multiple learnable domain experts (e.g., batch norm statistic, auxiliary classifiers, etc.) to capture domain-specific knowledge from different source domains individually. Still, two significant challenges remain. First, previous work only exploits some weak experts, like the batch norm, to capture knowledge, which naturally hampers the capability of capturing essential domain-specific knowledge. Second, previous methods such as \cite{doprompt} focused on learning domain knowledge independently while overlooking the rich cross-domain information that all domain experts can contribute collectively for the target domain prediction.

To overcome the above problems, we propose an environment-aware prompt vision transformer (EPVT) for domain generalization of skin lesion recognition. On the one hand, inspired by the emerging prompt learning techniques that embed prompts into a model for adaptation to diverse downstream tasks \cite{mtl,vpt,vlpt}, we construct different prompt vectors to strengthen the learning of domain-specific knowledge for adaptation to diverse domains. Then, the self-attention mechanism of the vision transformer (ViT) \cite{vit16} is adopted to fully model the relationship between image tokens and prompt vectors. On the other hand, to encourage cross-domain information sharing while preserving the domain-specific knowledge of each domain prompt, we propose a domain prompt generator based on low-rank weights updating. The prompt generator enables multiple domain prompts to work collaboratively and benefit from each other for generalization to unknown domains. Additionally, we devise a domain mixup strategy to resolve the problem of co-occurring artifacts in dermoscopic images and mitigate the resulting noisy domain label assignments.

Our contributions can be summarized as: (1) We resolve an artifacts-derived biasing problem in skin cancer diagnosis using a novel environment-aware prompt learning-based DG algorithm, EPVT; (2) EPVT takes advantage of a ViT-based domain-aware prompt learning and a novel domain prompt generator to improve domain-specific and cross-domain knowledge learning simultaneously; (3) A domain mixup strategy is devised to reduce the co-artifacts specific to dermoscopic images; (4) Extensive experiments on four out-of-distribution skin datasets and six biased ISIC datasets demonstrate the outperforming generalization ability and robustness of EPVT under heterogeneous distribution shifts.

\begin{figure*}[!t]
   \begin{center}
   {\includegraphics[width=1\linewidth]{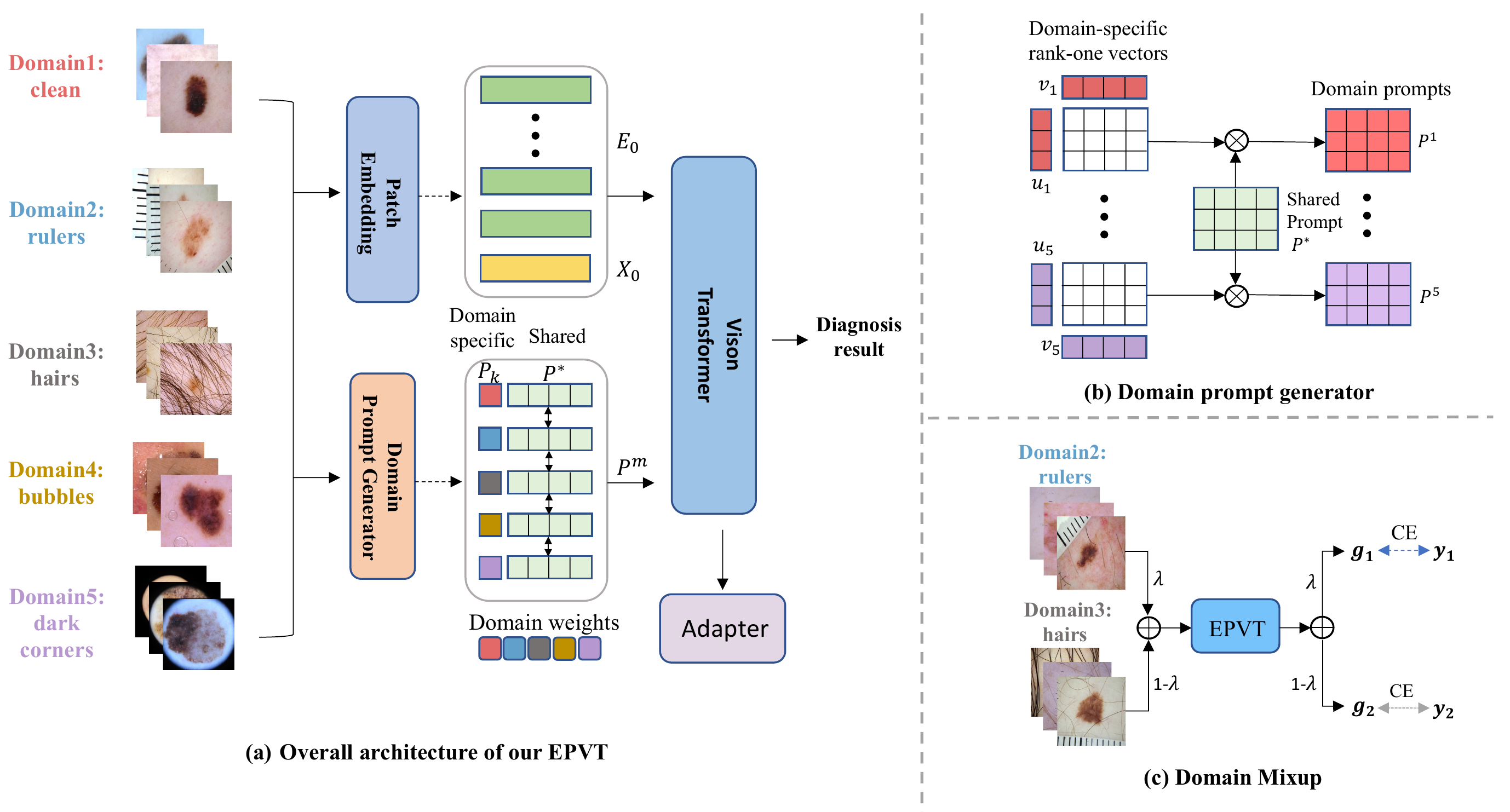}}
   \end{center}
  \vspace{-5mm}
\caption{\small 
The overview of our environment-aware prompt vision transformer (EPVT).}
   \label{fig:overview}
\end{figure*}
\section{Method}
 In domain generalization (DG), the training dataset $D_{train}$ consists of $M$ source domains, denoted as $D_{train}=\lbrace D_{k}| k=1,...,M\rbrace$. Here, each source domain $D_{k}$ is represented by $n$ labeled instances $\lbrace(x_{j}^{k},y_{j}^{k})\rbrace_{j=1}^{n}$. The goal of DG is to learn a model $G: X\rightarrow Y$ from the $M$ source domains so that it can generalize well in unseen target domains $D_{test}$. The overall architecture of our proposed model, EPVT, is shown in Fig.\ref{fig:overview}.a. We will illustrate its details in the following sections.
\subsection{Domain-specific Prompt Learning with Vision Transformer}
To enable the pre-trained vision transformer (ViT) to capture knowledge from different domains, as shown in Figure \ref{fig:overview}.a, we define a set of $M$ learnable domain prompts produced by a domain prompt generator (introduced in section \ref{sec2.2}), denoted as $P_{D}=\lbrace P^{m} \in \mathbb{R}^{d} \rbrace_{m=1}^{M}$, where $d$ is the same size as the feature embedding of the ViT and each prompt $P^{m}$ corresponds to one domain (\ie~dark corners). To incorporate these prompts into the model, we follow the conventional practice of visual prompt tuning \cite{vpt}, which prepends the prompts $P_{D}$ into the first layer of the transformer. Particularly, for each prompt $P^{m}$ in $P_{D}$, we extract the domain-specific features as:
\begin{equation}
F_{m}(x) = F([\ X_{0}, P^{m}, E_{0}\ ])
\end{equation}
where $F$ is the feature encoder of the ViT, $X_{0}$ denotes the class token, $E_{0}$ is the image patch embedding, $F_{m}$ is the feature extracted by ViT with the $m$-th prompt, and $0$ is the index of the first layer. Domain prompts $P_{D}$ are a set of learnable tokens, with each prompt $P^{m}$ being fed into the vision transformer along with the image and corresponding class tokens from a specific domain. Through optimizing, each prompt becomes a domain expert only responsible for the images from its own domain. By the self-attention mechanism of ViT, the model can effectively capture domain-specific knowledge from the domain prompt tokens.
\subsection{Cross-domain Knowledge Learning}
\label{sec2.2}
To facilitate effective knowledge sharing across different domains while maintaining its own parameters of each domain prompt, we propose a domain prompt generator, as depicted in Figure \ref{fig:overview}.b. Our approach is inspired by model adaptation and multi-task learning techniques used in natural language processing \cite{mtl,compacter}. Aghajanyan \etal  \cite{low_rank} have shown that when adapting a model to a specific task, the updates to weights possess a low intrinsic rank. Similarly, each domain prompt $P^{m}$ should also have a unique low intrinsic rank when learning knowledge from its own domain. To this end, we decompose each $P^{m}$ into a Hadamard product between a randomly initialized shared prompt $P^{*}$ and a rank-one matrix $P_{k}$ obtained from two randomly initialized learnable vectors $u_{k}$ and $v_{k}$, which is:
\begin{equation}
P^{m} = P^{*} \odot P_{k} \quad \textrm{where} \quad P_{k} = u_{k} \cdot v_{k}^{T}
\end{equation}

where $P^{m}$ represents the domain-specific prompt, computed by Hadamard product of $P^{*}$ and $P_{k}$. Here, $P^{*} \in \mathbb{R}^{s \times d}$ is utilized to learn general knowledge, with $s$ and $d$ representing the dimensions of the prompt vector and feature embedding respectively. On the other hand, $P_{k}$ is computed using domain-specific trainable vectors: $u_{k} \in \mathbb{R}^{s}$ and $v_{k} \in \mathbb{R}^{d}$. These vectors capture domain-specific information in a low-rank space. The decomposition of domain prompts into rank-one subspaces ensures that the model effectively encodes domain-specific information. By using the Hadamard product, the model can efficiently leverage cross-domain knowledge for target domain prediction.

\subsection{Mitigating the Co-artifacts Issue}
The artifacts-based domain labels can provide domain information for dermoscopic images. However, a non-trivial issue arises due to the possible co-occurrence of different artifacts from other domains within each domain. To address this issue, we employ a domain mixup strategy \cite{mixup1,mixup2}. Instead of assigning a hard prediction label ("0" or "1") to each image, in each batch, we mix every image using two randomly selected images from two different domains. This allows us to learn a flexible margin relative to both domains. We then apply the cross-entropy loss to the corresponding labels of bot images, as shown in \ref{fig:overview}.c and can be represented by the following equation:
\begin{equation}
\mathcal{L}_{mixup}=\lambda \mathcal{L}_{CE}(G(x_{mix}),y_{i})+(1-\lambda) \mathcal{L}_{CE}(G(x_{mix}),y_{j})
\label{mixup}
\end{equation}
where $x_{mix}=\lambda x_{i}^{k}+(1-\lambda)x_{j}^{q}$; $x_{i}^{k}$ and $x_{j}^{q}$ are samples from two different domains $k$ and $q$, and $y_{i}^{k}$ and $y_{j}^{q}$ are the corresponding labels. This strategy can overcome the challenge of ambiguous domain labels in dermoscopic images and improve the performance of our model.

\subsection{Optimization}
So far, we have introduced $\mathcal{L}_{mixup}$ in Eq. \ref{mixup} for optimizing our model. However, since our goal is to generalize the model to unseen environments, we also need to take advantage of each domain prompt. Instead of assigning equal weights to each domain prompt, we employ an adapter \cite{doprompt} that learns the linear correlation between the domain prompts and the target image prediction.
To obtain the adapted prompt for inference in the target domain, we define it as a linear combination of the source domain prompts:
\begin{equation}
P_{adapted}=A(F(x))=\sum_{m=1}^{M}w_{m}\cdot P^{m} \textrm{,} \quad \textrm{s.t.} \quad \sum_{m=1}^{M}w_{m}=1
\end{equation}
where $A$ represents an adapter containing a two-layer MLP with a softmax layer, and $w_m$ denotes the learned weights. 

To train the adapter $A$, we simulate the inference process for each image in the source domain by treating it as an image from the pseudo-target domain. Specifically, we first extract features from the ViT: $\hat{F_{m}}(x)=F([X_{0},E_{0}])$. Then we calculated the adapted prompt $P_{adapted}$ for the pseudo-target environment image x using the adapter $A$: $P_{adapted}=A(\hat{F_{m}}(x))$. Next, we extract features from ViT using the adapted prompt: $\hat{F_{m}}(x)=F([\hat{F_{m}}(x),P_{adapted},E_{0}])$. Finally, the classification head $H$ is applied to predict the label y: $y=H(\hat{F}_{m}(x))$. Additionally, the inferece process is the same as the simulated inference process and our final prediction will be conditioned on the adapted prompt $P_{adapted}$. 

To ensure that the adapter learns the correct linear correlation between the domain prompts and the target image, we use the domain label from source domains to directly supervise the weights $w_{m}$. We also use the cross-entropy loss to maintain the model performance with the adapted prompt:
\begin{equation}
\mathcal{L}_{adapted}=\mathcal{L}_{CE}(H(\hat{F}_{m}(x)),y)+ \lambda (\frac{1}{M}\sum_{m=1}^{M}\frac{1}{M}(\mathcal{L}_{CE}(w_{m}^{m},1)+\sum_{t\neq m}\mathcal{L}_{CE}(w_{t}^{m},0))
\label{eq9}
\end{equation}
where $\hat{F}_{m}(x)$ is the obtained feature map conditioned on the adapted prompt $P_{adapted}$, and $H$ is the classification head. The total loss is then defined as $\mathcal{L}_{total}=\mathcal{L}_{mixup}+\mathcal{L}_{adapted}$.
\section{Experiments}
\textbf{Experimental Setup}: We consider two challenging melanoma-benign classification settings that can effectively evaluate the generalization ability of our model in different environments and closely mimic real-world scenarios. (1) Out-of-distribution evaluation: The task is to evaluate the model on test sets that contain different artifacts or attributes compared to the training set. We train and validate all algorithms on \textit{ISIC2019} \cite{isic} dataset, following the split of \cite{artifacts}. We use the artifacts annotations from \cite{artifacts} and divide the training set of \textit{ISIC2019} into five groups: \textit{dark corner}, \textit{hair}, \textit{gel bubble}, \textit{ruler}, and \textit{clean}, with 2351, 4884, 1640, 672, and 2796 images, respectively. We evaluate models on four out-of-distribution (OOD) datasets, including \textit{Derm7pt-Dermoscopic} \cite{derm7pt}, \textit{Derm7pt-Clinical} \cite{derm7pt}, \textit{PH2} \cite{ph2}, and \textit{PAD-UFES-20} \cite{pad}. It's worth noting that \textit{ISIC2019}, \textit{Derm7pt-Dermoscopic}, and \textit{PH2} are dermoscopic images, while \textit{Derm7pt-Clinical} and \textit{PAD} are clinical images. (2) Trap set debiasing: We train and test our EPVT with its baseline on six trap sets \cite{artifacts} with increasing bias levels, ranging from 0 (randomly split training and testing sets from the ISIC2019 dataset) to 1 (the highest bias level where the correlation between artifacts and class label is in the opposite direction in the dataset splits). More details about these datasets and splits are provided in the complementary material. \vspace{1em}\\
\noindent\textbf{Implementation Details}: For a fair comparison, we train all models using ViT-Base/16 \cite{vit16} backbone pre-trained on Imagenet and report the ROC-AUC with five random seeds. Hyperparameter and model selection methods are crucial for domain generalization algorithms. We conduct a grid search over learning rate (from $3e^{-4}$ to $5e^{-6}$), weight decay (from $1e^{-2}$ to $1e^{-5}$), and the length of the prompt (from 4 to 16, when available) and report the best performance of all models. We employ the training-domain validation set method \cite{dg} for model selection. After the grid search, we use the AdamW optimizer with a learning rate of $5e^{-6}$ and a weight decay of $1e^{-2}$. The batch size is 130, and the length of the prompt is 10. We resize the input image to a size of $224\times224$ and adopt the standard data augmentation like random flip, crop, rotation, and color jitter. An early stopping with the patience of 22 is set and with a total of 60 epochs for OOD evaluation and 100 epochs for trap set debiasing. All experiments are conducted on a single NVIDIA RTX 3090 GPU.\vspace{1em}\\
\noindent\textbf{Out-of-distribution Evaluation}: Table 1 presents a comprehensive comparison of our EPVT algorithm with existing domain generalization methods. The results clearly demonstrate the superiority of our approach, with the best performance on three out of four OOD datasets and remarkable improvements over the ERM algorithm, especially achieving 4.1\% and 8.9\% improvement on the \textit{PAD} and \textit{PH2} datasets, respectively. Although some algorithms may perform similarly to our model on one of the four datasets, none can consistently match the performance of our method across all four datasets. Particularly, our approach showcases the highest average performance, with a 2.05\% improvement over the second-best algorithm across all four datasets. These findings highlight the effectiveness of our algorithm in learning robust features and its strong generalization abilities across diverse environments.\\

\begin{table}[!t]
\centering
\caption{The comparison on out-of-distribution datasets}
\label{tab1}
{\begin{tabular}{p{2.1cm}|p{1.9cm}|p{1.9cm}|p{1.9cm}|p{1.9cm}|p{1.9cm}}

\hline
Method & derm7pt\_d & derm7pt\_c & pad & ph2 & Average \\ \hline 
ERM  &$81.24 \pm  1.6$& $71.61 \pm 1.9$ & $82.62 \pm 1.6$ & $83.06 \pm 1.9$ & $79.63\pm1.5$ \\ 
DRO \cite{dro} & $\underline{82.46}\pm1.7$ & $72.88\pm1.9$ & $81.52\pm1.2$ & $84.64\pm1.8$ & $81.27\pm1.6$ \\  
CORAL \cite{coral} & $81.42\pm1.9$ & $71.45\pm1.3$ & $\textbf{88.13} \pm 1.2$ & $85.2\pm2.2$ & $81.55\pm1.5$ \\ 
MMD \cite{inv2} & $82.08\pm1.7$ & $71.8\pm1.5$ & $85.89\pm 1.9$ & $87.17\pm1.4$& $81.73\pm1.5$ \\ 
DANN\cite{dann} & $81.79\pm1.1$ & $73.12\pm1.6$ & $84.12\pm1.6$ & $85.18\pm1.9$ & $81.87\pm1.7$ \\  
IRM \cite{IRM} & $79.07\pm1.7$ & $71.3\pm1.8$ & $77.82\pm3.4$ & $79.37\pm1.2$ & $76.64\pm1.7$ \\ 
SagNet \cite{sagnet} & $82.28\pm1.8$ & $73.19\pm1.6$ & $78.89\pm 4.5$ & $88.79\pm1.9$ & $81.79\pm1.8$ \\ 

MLDG\cite{meta}& $81.06\pm1.6$ & $71.79\pm1.6$ & $83.41\pm1.0$ & $84.22\pm1.8$ & $79.87\pm1.2$ \\ 
CAD \cite{cad} & $82.72\pm1.5$ & $69.57\pm1.6$ & $81.36\pm1.9$ & $88.4\pm1.5$ & $81.51\pm1.6$ \\ 
DoPrompt \cite{doprompt} & $82.38\pm1.0$ & $71.61\pm1.7$& $83.81\pm1.4$& $\underline{91.33}\pm1.8$ & $\underline{82.06}\pm1.6$ \\ 
SelfReg \cite{selfreg}& $81.83\pm1.9$ & $\underline{73.29}\pm1.4$ & $85.27\pm1.3$ & $85.16\pm3.3$ & $81.12\pm1.0$\\  \hline
EPVT (Ours) & $\textbf{83.69}\pm1.4$ &$\textbf{73.96}\pm1.6$ & $\underline{86.67}\pm1.5$& $\textbf{91.91}\pm1.5$& $\textbf{84.11}\pm1.4$\\ 
\hline
\end{tabular}
}
\end{table}

\begin{table}[!t]
\centering
\caption{Ablation study on out-of-distribution datasets}
\label{tab2}
{\begin{tabular}{p{2.1cm}|p{1.9cm}|p{1.9cm}|p{1.9cm}|p{1.9cm}|p{1.9cm}}
\hline
Method & derm7pt\_d & derm7pt\_c & pad & ph2 & Average \\ \hline 
Baseline   &$81.24 \pm  1.6$& $71.61 \pm 1.9$ & $82.62 \pm 1.6$ & $83.06 \pm 1.9$ & $79.63\pm1.5$ \\  
+P  & $82.13\pm1.1$ & $71.41\pm1.3$ & $82.15\pm1.6$ & $84.21\pm1.4$ & $79.73\pm1.3$ \\ 
+P+A  & $82.55\pm1.6$ & $72.86\pm1.1$ & $81.02\pm1.5$ & $84.97\pm1.8$ & $81.10\pm1.6$ \\  
+P+A+M& $81.43\pm1.4$ & $73.18\pm1.5$ & $85.78\pm1.9$ & $89.28\pm1.3$ & $82.42\pm1.7$ \\  
+P+A+M+G  & $\textbf{83.69}\pm1.4$ &$\textbf{73.96}\pm1.6$ & $\textbf{86.67}\pm1.5$& $\textbf{91.91}\pm1.5$& $\textbf{84.11}\pm1.4$\\  \hline
\end{tabular}%
}
\end{table}
\noindent\textbf{Ablation Study}: We perform ablation studies to analyze each component of our model, as shown in Table \ref{tab2}. We set our baseline as the Empirical Risk Minimization (ERM) algorithm, and we gradually add P (prompt \cite{vpt}), A (Adapter), M (Mixup), and G (domain prompt generator) into the model. Firstly, we observe that the baseline model with prompt only improves the average performance by 0.1\%, showing that simply combining prompt does not very helpful for domain generalization. When we combine the adapter, the model's average performance improves by 1.37\%, but it performs worse than ERM on \textit{PAD} dataset. Subsequently, we added domain mixup and domain prompt generator to the model, resulting in significant further improvements in the model's average performance by 1.32\% and 1.69\%, respectively. The consistently better performance than the baseline on all four datasets also highlights the importance of addressing co-artifacts and cross-domain learning for DG in skin lesion recognition.\\

\begin{figure}[!t]
    \centering
\subfigure{\includegraphics[width=1.0\textwidth]{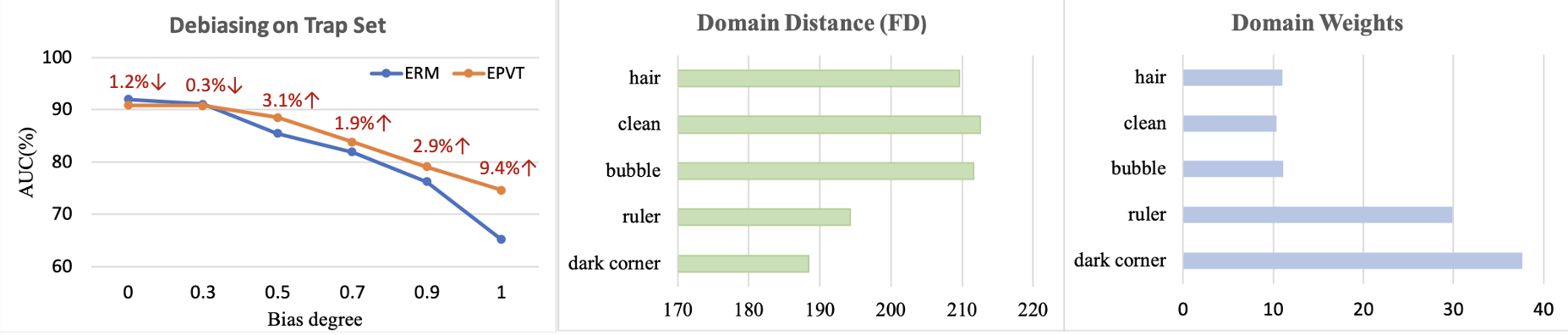}} 
  \vspace{-5mm}
    \caption{(a)Deibiasing evaluation (b) domain distance (c) domain weights }
    \label{fig2}
\end{figure}
\noindent\textbf{Trap Set Debiasing}: In Figure \ref{fig2}.a, we present the performance of the ERM baseline and our EPVT on six biased ISIC2019 datasets. Each point on the graph represents an algorithm that is trained and tested on a specific bias degree split. The graph shows that the ERM baseline performs better than our EPVT when the bias is low (0 and 0.3). However, this is because ERM relies heavily on spurious correlations between artifacts and class labels, leading to overfitting on the training set. As the bias degree increases, the correlation between artifacts and class labels decreases, and overfitting the train set causes the performance of ERM to drop dramatically on the test set with a significant distribution difference. In contrast, our EPVT exhibits greater robustness to different bias levels. Notably, our EPVT outperforms the ERM baseline by 9.4\% on the bias 1 dataset.\\

\noindent\textbf{Prompt Weights Analysis}: To verify whether our model has learned the correct domain prompts for target domain prediction, we analyze and plot the results in Figure \ref{fig2}.b and \ref{fig2}.c. Firstly, we extract the features of each domain from our training set and extract the feature from one target dataset, \textit{Derm7pt-Clin}. We then calculate the Frechet distance \cite{frechet} between each domain and the target dataset using the extracted feature, representing the domain distance between them. The results are recorded in Figure \ref{fig2}.b. Next, we record the learned weights of each domain prompt in Figure \ref{fig2}.c; it shows that our model assigns the highest weight to the "dark corner" group, as the domain distance between "dark corner" and \textit{Derm7pt-Clin} is the closest, as shown in Figure \ref{fig2}.b. This suggests that they share the most similar domain information. Further, the "clean" group is assigned the smallest weight as the domain distance between them is the largest, indicating that their domains are significantly different and contain less useful information for target domain prediction. In summary, we observe a negative correlation between domain distance and the prompt's weights, indicating that our model can learn the correct knowledge from different domains precisely.

\section{Conclusion}
In this paper, we propose a novel DG algorithm called EPVT for robust skin lesion recognition. Our approach addresses the co-artifacts problem using a domain mixup strategy and cross-domain learning problems using a domain prompt generator. Compared to other competitive domain generalization algorithms, our method achieves outstanding results on three out of four OOD datasets and the second-best on the remaining one. Additionally, we conducted a debiasing experiment that highlights the shortcomings of conventional training using empirical risk minimization, which leads to overfitting in dermoscopic images due to artifacts. In contrast, our EPVT model effectively reduces overfitting and consistently performs better in different biased environments.
\bibliographystyle{splncs04}
\bibliography{arxiv}

\begin{table}[h]
\centering
\caption{Datasets statistic.}
\begin{tabular}{p{3.5cm}p{2cm}p{2cm}p{2cm}}
\hline
Dataset          & No. of Image & Split & Type        \\ \hline
ISIC2019 (train)  & 12360        & train & dermoscopic \\ 
ISIC2019 (val)   & 2057         & val   & dermoscopic \\ 
ISIC2019 (test) & 6199         & test  & dermoscopic \\ 
Derm7pt\_Derm    & 872          & test  & dermoscopic \\ 
Derm7pt\_Clin    & 839          & test  & clinical    \\ 
PAD-UFES-20   & 531          & test  & clinical    \\ 
PH2   & 200         & test  & dermoscopic \\ \hline
\end{tabular}
\end{table}

\begin{figure}[h]
    \centering
\subfigure{\includegraphics[width=0.8\textwidth]{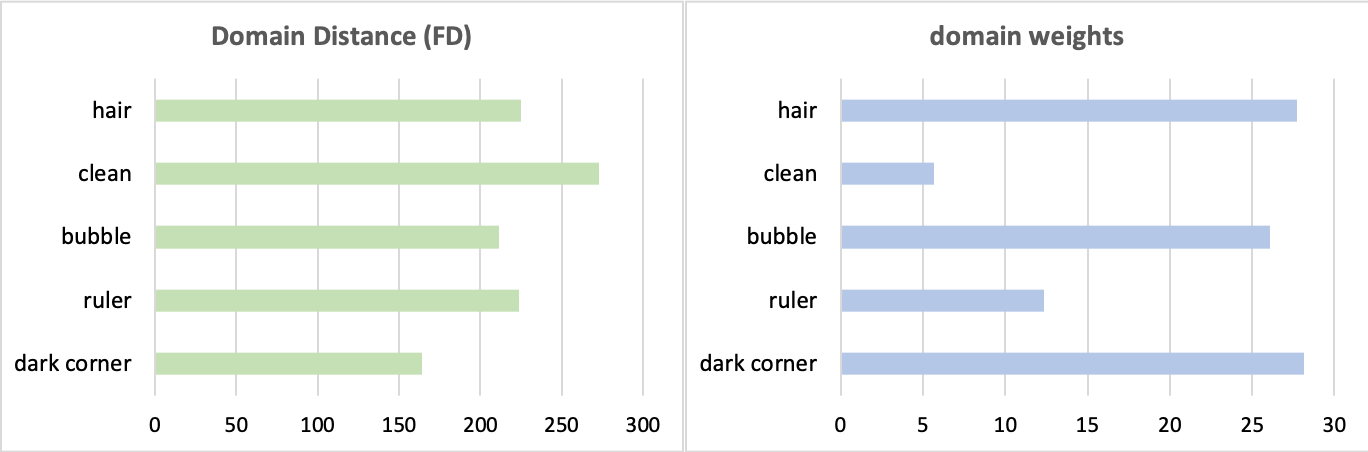}} 
  \vspace{-5mm}
    \caption{domain ditance vs prompt weights on PH2 }
    \label{fig2}
\end{figure}

\begin{table}[h]
\centering
\caption{Domain distance between the source domain and target domains.}
\begin{tabular}{p{2.8cm}p{2cm}p{2cm}p{2cm}p{2cm}}
\hline
Domain distance    & Derm7pt\_D& Derm7pt\_C & PAD&PH2        \\ \hline
ISIC2019 (train) & 122.82  & 216.4 & 187.85&300.47 \\ \hline
\end{tabular}
\end{table}
\begin{figure*}[!t]
   \centering
   \renewcommand{\arraystretch}{0.2}
   \begin{tabular}{ c c c c c}
      \resizebox{0.18\linewidth}{!}{\includegraphics{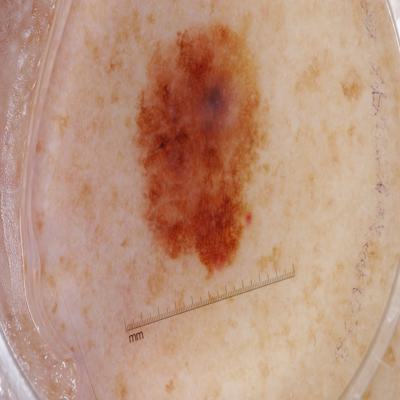}} &
        \resizebox{0.18\linewidth}{!}{\includegraphics{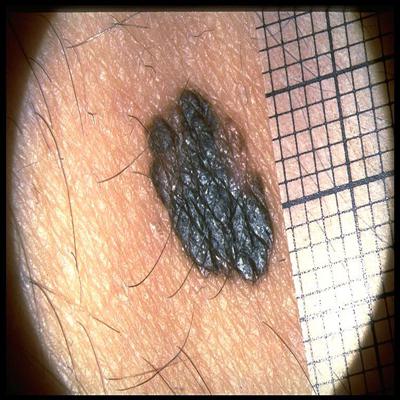}} &
      \resizebox{0.18\linewidth}{!}{\includegraphics{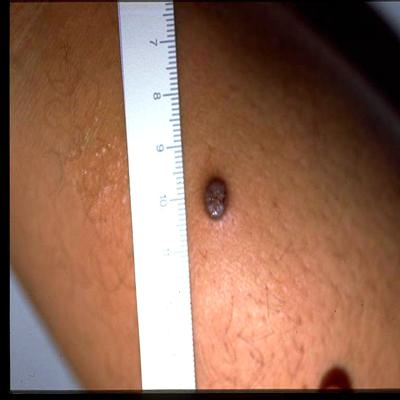}} &
      \resizebox{0.18\linewidth}{!}{\includegraphics{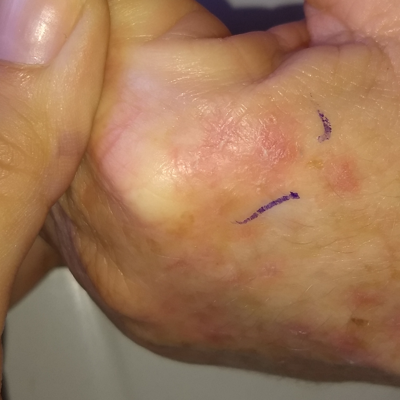}} &
      \resizebox{0.18\linewidth}{!}{\includegraphics{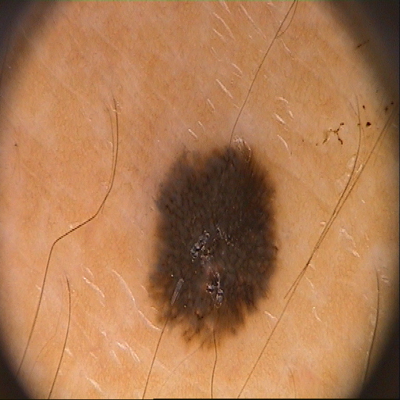}} \\
      \resizebox{0.18\linewidth}{!}{\includegraphics{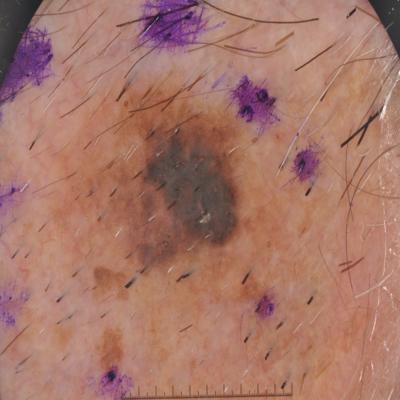}} &
        \resizebox{0.18\linewidth}{!}{\includegraphics{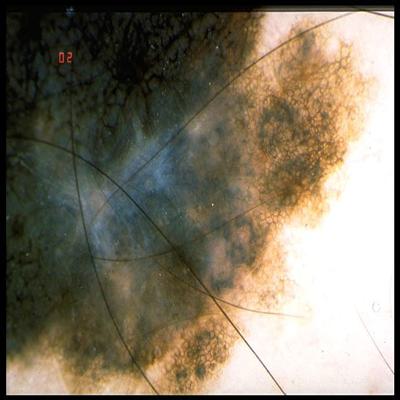}} &
      \resizebox{0.18\linewidth}{!}{\includegraphics{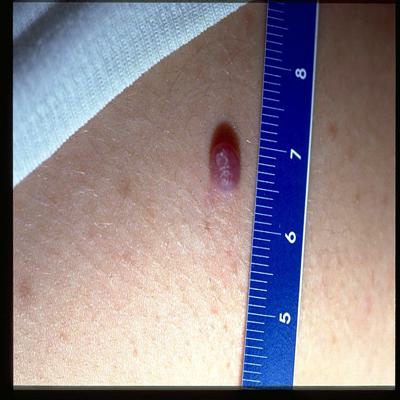}} &
      \resizebox{0.18\linewidth}{!}{\includegraphics{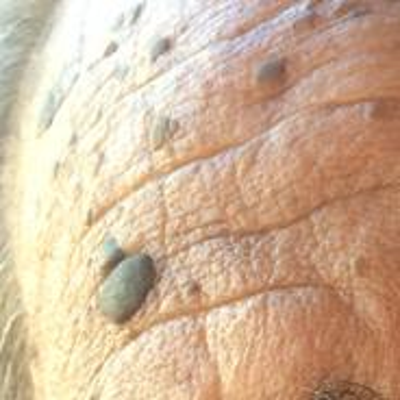}} &
      \resizebox{0.18\linewidth}{!}{\includegraphics{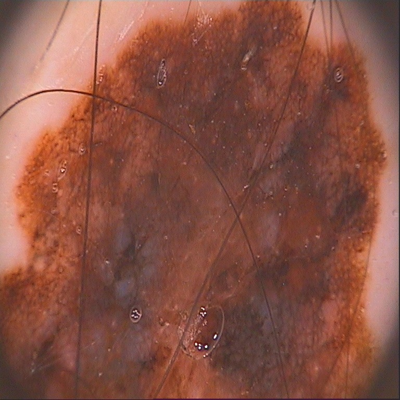}} \\
      \resizebox{0.18\linewidth}{!}{\includegraphics{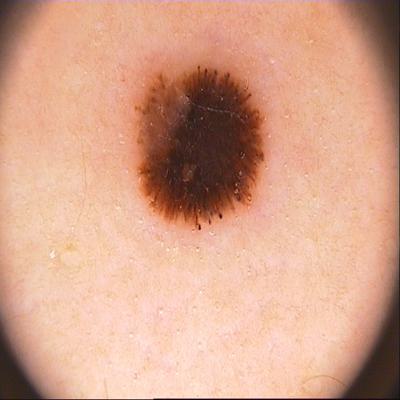}} &
        \resizebox{0.18\linewidth}{!}{\includegraphics{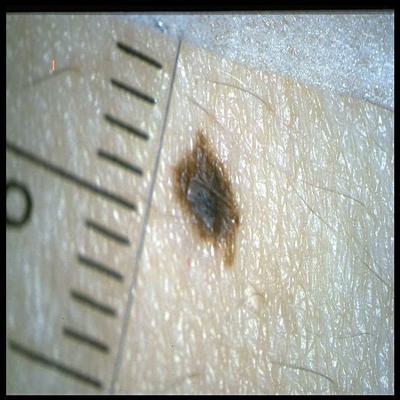}} &
      \resizebox{0.18\linewidth}{!}{\includegraphics{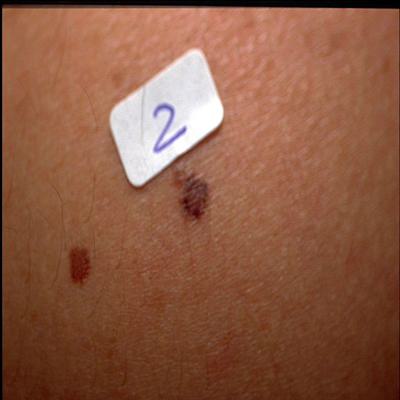}} &
      \resizebox{0.18\linewidth}{!}{\includegraphics{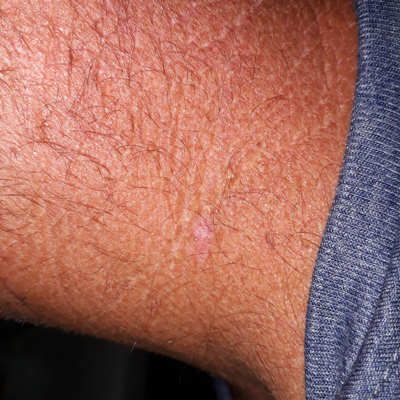}} &
      \resizebox{0.18\linewidth}{!}{\includegraphics{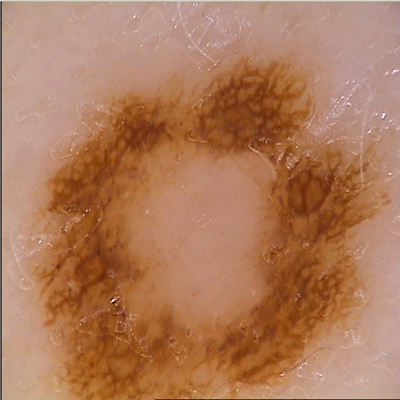}} \\
          \resizebox{0.18\linewidth}{!}{\includegraphics{supp_figure/isic4.jpg}} &
        \resizebox{0.18\linewidth}{!}{\includegraphics{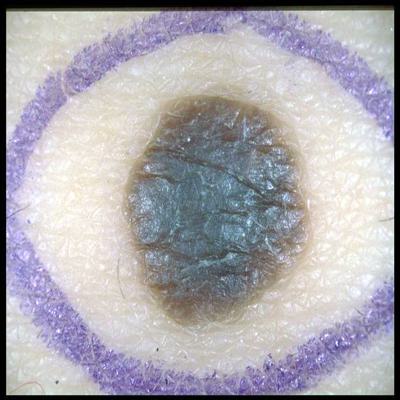}} &
      \resizebox{0.18\linewidth}{!}{\includegraphics{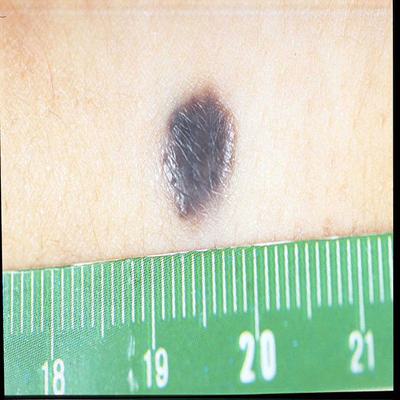}} &
      \resizebox{0.18\linewidth}{!}{\includegraphics{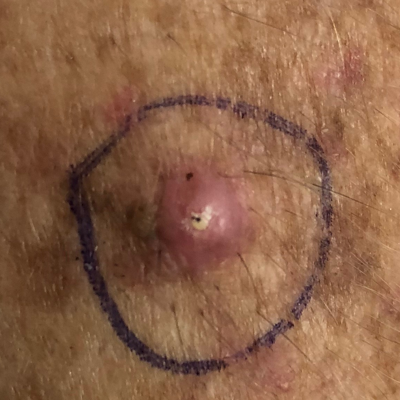}} &
      \resizebox{0.18\linewidth}{!}{\includegraphics{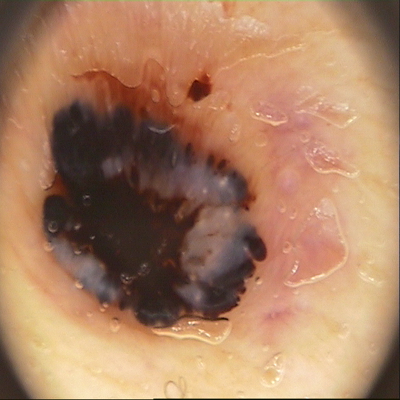}} \\
          \resizebox{0.18\linewidth}{!}{\includegraphics{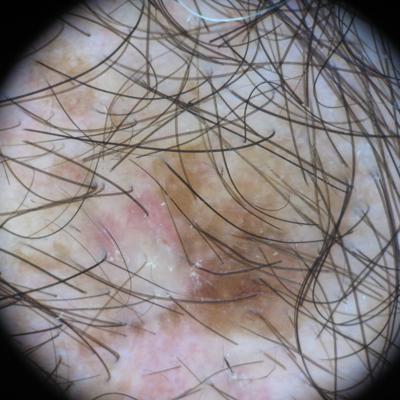}} &
        \resizebox{0.18\linewidth}{!}{\includegraphics{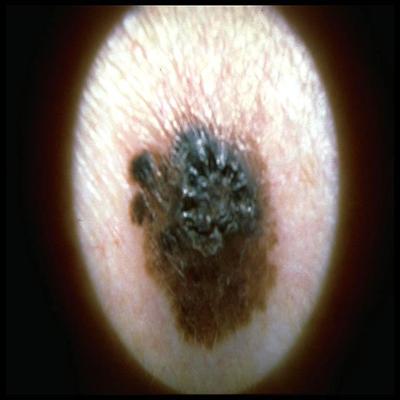}} &
      \resizebox{0.18\linewidth}{!}{\includegraphics{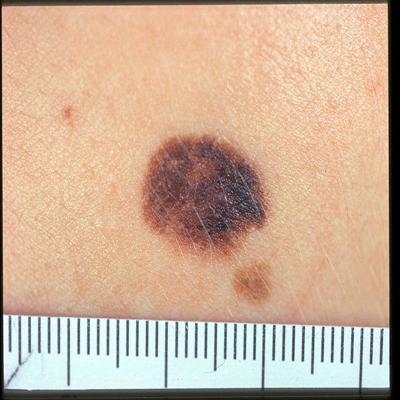}} &
      \resizebox{0.18\linewidth}{!}{\includegraphics{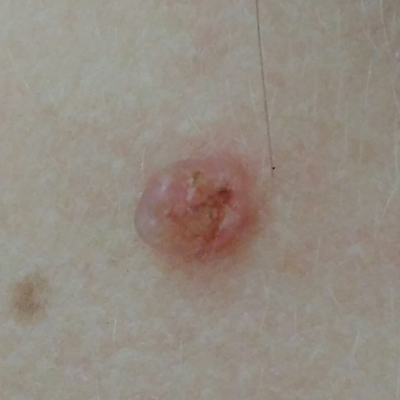}} &
      \resizebox{0.18\linewidth}{!}{\includegraphics{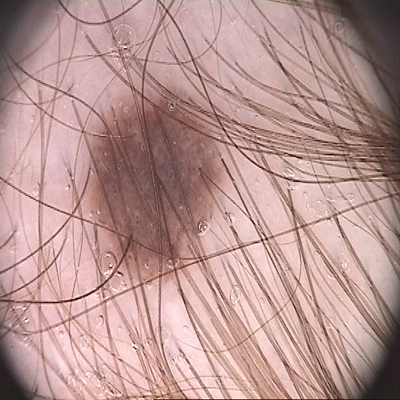}} \\
   \\
   \footnotesize{ISIC2019} & \footnotesize{Derm7pt\_D} & \footnotesize{Derm7pt\_C} & \footnotesize{PAD} & \footnotesize{PH2}  \\
   \end{tabular}
   \caption{Example images from each dataset.}
   \label{fig:example}
\end{figure*}

\end{document}